\documentclass[10pt,twocolumn,letterpaper]{article} 

\usepackage{cvpr}
\usepackage{times}
\usepackage{epsfig}
\usepackage{graphicx}
\usepackage{booktabs}
\usepackage{comment}
\usepackage{amsmath}
\usepackage{amssymb}
\usepackage{enumitem}

\usepackage{caption}
\usepackage{subcaption}
\usepackage{icomma}
\usepackage{setspace}
\usepackage{multirow}
\usepackage{balance}
\newcommand{\rama}{\textcolor{black}}
\newcommand{\satwik}{\textcolor{black}}

\newcommand{\vwv}{\texttt{vis-w2v}}
\newcommand{\wv}{\texttt{w2v}}
\newcommand{\wiki}{\texttt{-wiki}}
\newcommand{\coco}{\texttt{-coco}}
\newcommand{\reffig}[1]{Fig.~\ref{#1}}
\newcommand{\refsec}[1]{Sec.~\ref{#1}}
\newcommand{\refeq}[1]{Eq.~\ref{#1}}
\newcommand{\compactparagraph}[1]{\vspace*{5pt} \noindent \textbf{#1}}

\usepackage[breaklinks=true,bookmarks=false]{hyperref}

\cvprfinalcopy 

\def\cvprPaperID{0193} 

\ifcvprfinal\fi
\begin{document}

\title{Visual Word2Vec (\vwv{}): Learning Visually Grounded \\Word Embeddings Using Abstract Scenes}

\author{
Satwik Kottur$^{1\ast}$ \quad Ramakrishna Vedantam$^{2}$\thanks{Equal contribution} \quad Jos{\'e} M. F. Moura$^1$ \quad Devi Parikh$^2$\\
$^1$Carnegie Mellon University \quad $^2$Virginia Tech \\
{{\tt\small $^1$skottur@andrew.cmu.edu,moura@ece.cmu.edu} \quad {\tt\small$^2$\{vrama91,parikh\}@vt.edu}}
}
\maketitle

\begin{abstract}
	We propose a model to learn visually grounded word embeddings (\emph{\vwv{}}) to capture visual notions of semantic relatedness. While word embeddings trained using text have been extremely successful, they cannot uncover notions of semantic relatedness implicit in our visual world. For instance, although ``eats'' and ``stares at'' seem unrelated in text, they share semantics visually. When people are eating something, they also tend to stare at the food. Grounding diverse relations like ``eats'' and ``stares at'' into vision remains challenging, despite recent progress in vision. We note that the visual grounding of words depends on semantics, and not the literal pixels. We thus use abstract scenes created from clipart to provide the visual grounding. We find that the embeddings we learn capture fine-grained, visually grounded notions of semantic relatedness. We  show improvements over text-only word embeddings (word2vec) on three tasks: common-sense assertion classification, visual paraphrasing and text-based image retrieval. Our code and datasets are available online. 

\end{abstract}

\vspace{-10pt}
\section{Introduction}
\begin{figure}
\includegraphics[width=\columnwidth]{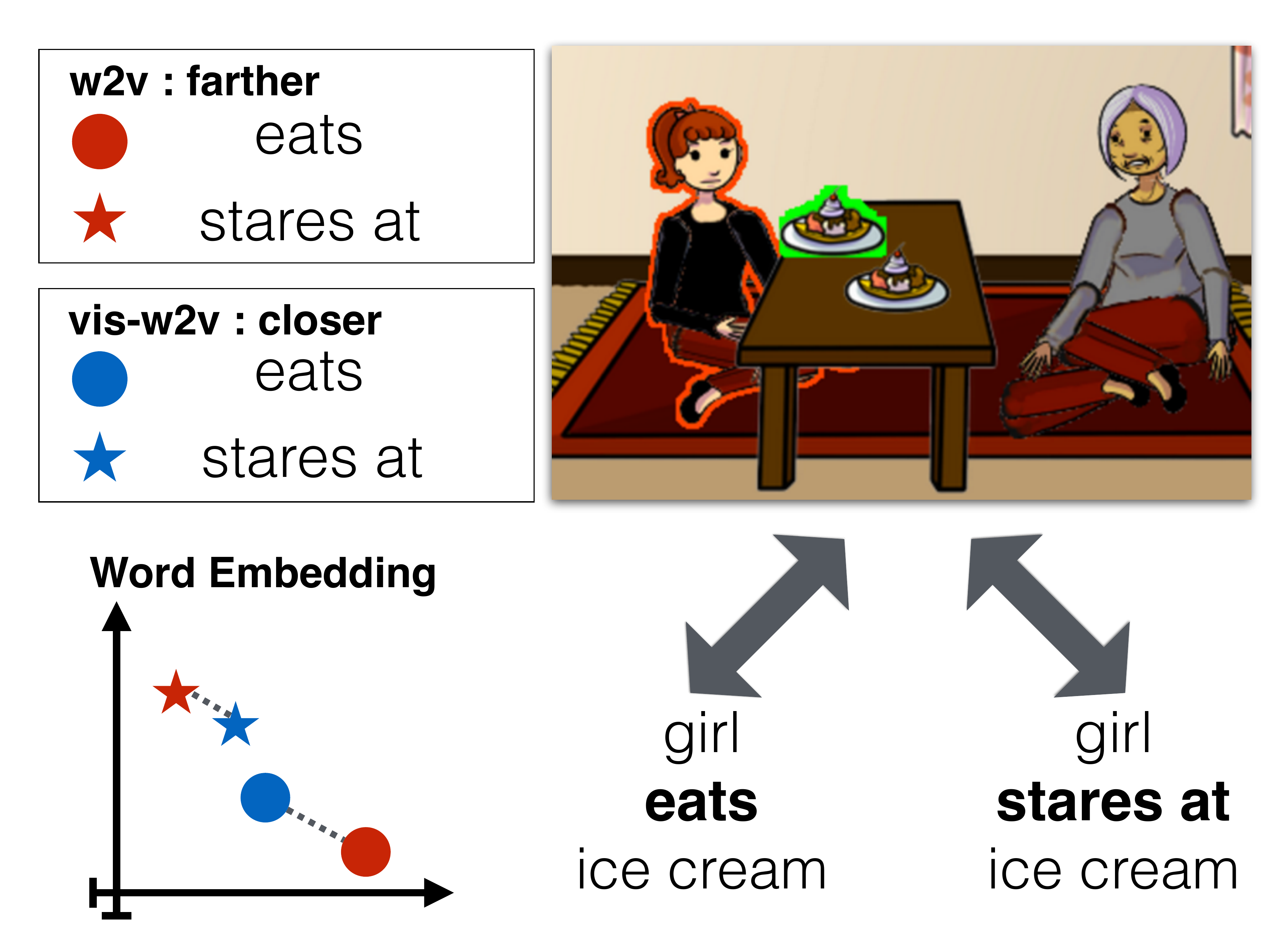}
\caption{We ground text-based word2vec (\wv{}) embeddings into vision to capture a complimentary notion of visual relatedness. Our method (\vwv{}) learns to predict the visual grounding as context for a given word. Although ``eats'' and ``stares at'' seem unrelated in text, they share semantics visually. Eating involves staring or looking at the food that is being eaten. As training proceeds, embeddings change from \wv{}~(red) to \vwv{}~(blue).}
\label{fig:teaser}
\vspace{-10pt}
\end{figure}

Artificial intelligence (AI) is an inherently multi-modal problem: understanding and reasoning about multiple modalities (as humans do), seems crucial for achieving artificial intelligence (AI). Language and vision are two vital interaction modalities for humans. Thus, modeling the rich interplay between language and vision is one of fundamental problems in AI.

Language modeling is an important problem in natural language processing (NLP). A language model estimates the likelihood of a word conditioned on other (context) words in a sentence. There is a rich history of works on $n$-gram based language modeling~\cite{Chen98anempirical,Katz87estimationof}. It has been shown that simple, count-based models trained on millions of sentences can give good results. However, in recent years, neural language models~\cite{Bengio03aneural,Mikolov2009} have been explored. Neural language models learn mappings ($W:words \rightarrow \mathbb{R}^n$) from words (encoded using a dictionary) to a real-valued vector space (embedding), to maximize the log-likelihood of words given context. Embedding words into such a vector space helps deal with the curse of dimensionality, so that we can reason about similarities between words more effectively. One popular architecture for learning such an embedding is word2vec~\cite{Mikolov2013a,Mikolov2013}. This embedding captures rich notions of semantic relatedness and compositionality between words~\cite{Mikolov2013}. 

For tasks at the intersection of vision and language, it seems prudent to model semantics as dictated by both text and vision. It is especially challenging to model fine-grained interactions between objects using only text. Consider the relations ``eats'' and ``stares at'' in \reffig{fig:teaser}. When reasoning using only text, it might prove difficult to realize that these relations are semantically similar. However, by grounding the concepts into vision, we can learn that these relations are more similar than indicated by text. Thus, visual grounding provides a complimentary notion of semantic relatedness. In this work, we learn word embeddings to capture this grounding.

Grounding fine-grained notions of semantic relatedness between words like ``eats'' and ``stares at'' into vision is a challenging problem. While recent years have seen tremendous progress in tasks like image classification~\cite{Krizhevsky2012ImageNetNetworks}, detection~\cite{girshick2014rcnn}, semantic segmentation~\cite{long_shelhamer_fcn}, action recognition~\cite{MajiActionCVPR11}, \etc., modeling fine-grained semantics of interactions between objects is still a challenging task. However, we observe that it is the semantics of the visual scene that matter for inferring the visually grounded semantic relatedness, and not the literal pixels (\reffig{fig:teaser}). We thus use abstract scenes made from clipart to provide the visual grounding. We show that the embeddings we learn using abstract scenes generalize to text describing real images (\refsec{subsec:cs_res}).

Our approach considers visual cues from abstract scenes as context for words. Given a set of words and associated abstract scenes, we first cluster the scenes in a rich semantic feature space capturing the presence and locations of objects, pose, expressions, gaze, age of people,~\etc. Note that these features can be trivially extracted from abstract scenes. Using these features helps us capture fine-grained notions of semantic relatedness (\reffig{fig:clustering}). We then train to predict the cluster membership from pre-initialized word embeddings. The idea is to bring embeddings for words with similar visual instantiations closer, and push words with different visual instantiations farther (\reffig{fig:teaser}). The word embeddings are initialized with word2vec~\cite{Mikolov2013}. The clusters thus act as surrogate classes. Note that each surrogate class may have images belonging to concepts which are different in text, but are visually similar. Since we predict the visual clusters as context given a set of input words, our model can be viewed as a multi-modal extension of the continuous bag of words (CBOW)~\cite{Mikolov2013} word2vec model.

\compactparagraph{Contributions:} We propose a novel model \emph{visual word2vec} (\vwv) to learn visually grounded word embeddings. We use abstract scenes made from clipart to provide the grounding. We demonstrate the benefit of \vwv~on three tasks which are ostensibly in text, but can benefit from visual grounding: common sense assertion classification~\cite{vedantamLICCV15}, visual paraphrasing~\cite{Lin_2015_CVPR}, \rama{and} text-based image retrieval~\cite{jas2015specificity}. Common sense assertion classification~\cite{vedantamLICCV15} is the task of modeling the plausibility of common sense assertions of the form (\texttt{boy}, \texttt{eats}, \texttt{cake}). Visual paraphrasing~\cite{Lin_2015_CVPR} is the task of determining whether two sentences describe the same underlying scene or not. Text-based image retrieval is the task of retrieving images by matching accompanying text with textual queries. We show consistent improvements over baseline word2vec (\wv) models on these tasks. Infact, on the common sense assertion classification task, our models surpass the state of the art.

The rest of the paper is organized as follows. \refsec{sec:related-work} discusses related work on learning word embeddings, learning from visual abstraction, \etc. \refsec{sec:approach} presents our approach. \refsec{sec:applications} describes the datasets we work with. We provide experimental details in \refsec{sec:experiments} and results in \refsec{sec:results}. 
\section{Related Work}\label{sec:related-work}
\compactparagraph{Word Embeddings:} Word embeddings learnt using neural networks~\cite{collobert:2008,Mikolov2013} have gained a lot of popularity recently. These embeddings are learnt offline and then typically used to initialize a multi-layer neural network language model~\cite{Bengio03aneural,Mikolov2009}. Similar to those approaches, we learn word embeddings from text offline, and finetune them to predict visual context. Xu~\etal~\cite{xuimproving} and Lazaridou~\etal~\cite{lazaridou-pham-baroni:2015:NAACL-HLT} use visual cues to improve the word2vec representation by predicting real image representations from word2vec and maximizing the dot product between image features and word2vec respectively. While their focus is on capturing appearance cues (separating cats and dogs based on different appearance), we instead focus on capturing fine-grained semantics using abstract scenes. We study if the model of Ren~\etal~\cite{xuimproving} and our \vwv{} provide complementary benefits in the appendix. Other works use visual and textual attributes (\eg vegetable is an attribute for potato) to improve distributional models of word meaning~\cite{conf/acl/SilbererFL13,silberer2014learning}. In contrast to these approaches, our set of visual concepts need not be explicitly specified, it is implicitly learnt in the clustering step. Many works use word embeddings as parts of larger models for tasks such as image retrieval~\cite{Kiros2014UnifyingModels}, image captioning~\cite{Kiros2014UnifyingModels,Vinyals2015ShowGenerator},~\etc. These multi-modal embeddings capture regularities like compositional structure between images and words. For instance, in such a multi-modal embedding space, ``image of blue car''~-~``blue"~+~``red'' would give a vector close to ``image of red car''. In contrast, we want to learn unimodal (textual) embeddings which capture multi-modal semantics. For example, we want to learn that ``eats'' and ``stares at'' are (visually) similar.

\compactparagraph{Surrogate Classification:} There has been a lot of recent work on learning with surrogate labels due to interest in unsupervised representation learning. Previous works have used surrogate labels to learn image features~\cite{doersch2015unsupervised,DB14b}. In contrast, we are interested in augmenting word embeddings with visual semantics. Also, while previous works have created surrogate labels using data transformations~\cite{DB14b} or sampling~\cite{doersch2015unsupervised}, we create surrogate labels by clustering abstract scenes in a semantically rich feature space. 

\compactparagraph{Learning from Visual Abstraction:} Visual abstractions have been used for a variety of high-level scene understanding tasks recently. Zitnick~\etal~\cite{Clipart_PAMI,Zitnick_2013_CVPR} learn the importance of various visual features (occurrence and co-occurrence of objects, expression, gaze, \etc) in determining the meaning or semantics of a scene.~\cite{Zitnick_2013_ICCV} and ~\cite{Fouhey_2014} learn the visual interpretation of sentences and the dynamics of objects in temporal abstract scenes respectively. Antol~\etal~\cite{Antol_2014} learn models of fine-grained interactions between pairs of people using visual abstractions. Lin and Parikh~\cite{Lin_2015_CVPR} ``imagine'' abstract scenes corresponding to text, and use the common sense depicted in these imagined scenes to solve textual tasks such as fill-in-the-blanks and paraphrasing. Vedantam~\etal~\cite{vedantamLICCV15} classify common sense assertions as plausible or not by using textual and visual cues. In this work, we experiment with the tasks of~\cite{Lin_2015_CVPR} and~\cite{vedantamLICCV15}, which are two tasks in text that could benefit from visual grounding. Interestingly, by learning \vwv{}, we eliminate the need for explicitly reasoning about abstract scenes at test time,~\ie, the visual grounding captured in our word embeddings suffices. 

\compactparagraph{Language, Vision and Common Sense:} There has been a surge of interest in problems at the intersection of language and vision recently. Breakthroughs have been made in tasks like image captioning~\cite{DBLP:journals/corr/ChenZ14a,DBLP:journals/corr/DonahueHGRVSD14,journals/jair/HodoshYH13,Karpathy2015DeepDescriptions,Kiros2014UnifyingModels,babytalk,DBLP:journals/corr/MaoXYWY14,midge,Vinyals2015ShowGenerator}, video description~\cite{DBLP:journals/corr/DonahueHGRVSD14,rohrbach13iccv}, visual question answering~\cite{VQA,gao2015mQA,Geman24032015,DBLP:journals/corr/MalinowskiF14,DBLP:journals/corr/MalinowskiRF15,DBLP:journals/corr/RenKZ15}, aligning text and vision~\cite{Karpathy2015DeepDescriptions,Kiros2014UnifyingModels},~\etc. In contrast to these tasks (which are all multi-modal), our tasks themselves are unimodal (\ie, in text), but benefit from using visual cues. Recent work has also studied how vision can help common sense reasoning~\cite{vedantamLICCV15,sadeghi2015viske}. In comparison to these works, our approach is generic,~\ie, can be used for multiple tasks (not just common sense reasoning).

\section{Approach}\label{sec:approach}
Recall that our \vwv{} model grounds word embeddings into vision by treating vision as context. We first detail our inputs. We then discuss our \vwv{} model. We then describe the clustering procedure to get surrogate semantic labels, which are used as visual context by our model. We then describe how word-embeddings are initialized. Finally, we draw connections to word2vec (\wv{}) models.
\begin{figure}[t]
	\centering
	\includegraphics[width=\columnwidth]{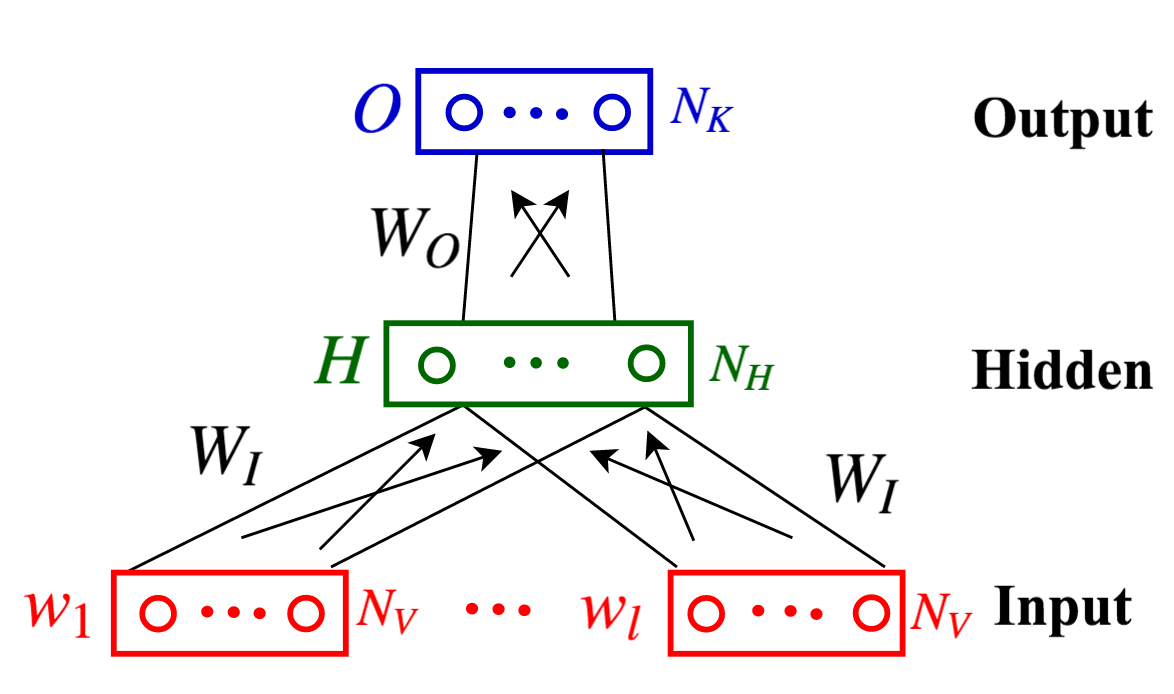}
    \caption{\satwik{Proposed \vwv{} model. The input layer (red) has multiple one-hot word encodings. These are connected to the hidden layer with the projection matrix $W_I$, \ie, all the inputs share the same weights. It is finally connected to the output layer via $W_O$. Model predicts the visual context $O$ given the text input $S_{w}=\{w_{l}\}$.}}
	\label{fig:vis-w2v-model}
    \vspace{-14pt}
\end{figure}

\compactparagraph{Input:} We are given a set of pairs of visual scenes and associated text $D = \{(v, w)\}_d$ in order to train \vwv{}. Here $v$ refers to the image features and $w$ refers to the set of words associated with the image. At each step of training, we select a window $S_{w} \subseteq w$ to train the model.

\compactparagraph{Model:}\label{subsec:model} Our \vwv{} model (\reffig{fig:vis-w2v-model}) is a neural network that accepts as input a set of words $S_{w}$ and a visual feature instance $v$. Each of the words $w_{i} \in S_{w}$ is represented via a one-hot encoding. A one-hot encoding enumerates over the set of words in a vocabulary (of size $N_V$) and places a 1 at the index corresponding to the given word. This one-hot encoded input is transformed using a projection matrix $W_{I}$ of size ${N_V\times N_H}$ that connects the input layer to the hidden layer, where the hidden layer has a dimension of $N_H$. Intuitively, $N_H$ decides the capacity of the representation. Consider an input one-hot encoded word $w_{i}$ whose $j^{th}$ index is set to 1. Since $w_{i}$ is one-hot encoded, the hidden activation for this word ($H_{w_{i}}$) is a row in the weight matrix $W_I^{j}$, \satwik{\ie, $H_{w_{i}} = W_I^{j}$.}
\satwik{The resultant hidden activation $H$ would then be the average of individual hidden activations $H_{w_i}$ as $W_I$ is shared among all the words $S_w$, \ie,:}
\begin{align}
    H &= \frac{1}{|S_{w}|} {\sum_{w_{i} \in S_{w} \subseteq w} {H_{w_{i}}}}	
\end{align}

Given the hidden activation $H$, we multiply it with an output weight matrix $W_{O}$ of size ${N_H\times N_K}$, where $N_K$ is the number of output classes. The output class (described next) is a discrete-valued function of the visual features $G(v)$ (more details in next paragraph). We normalize the output activations $O = H \times W_{O}$ to form a distribution using the softmax function. Given the softmax outputs, we minimize the negative log-likelihood of the correct class conditioned on the input words:
\begin{equation}\label{eqn:opt}
\min_{W_{I}, W_{O}} - \log{P(G(v)|S_{w}, W_{I}, W_{O})}
\end{equation}
We optimize for this objective using stochastic gradient descent (SGD) with a learning rate of 0.01.

\compactparagraph{Output Classes:}\label{sec:cluster} As mentioned in the previous section, the target classes for the neural network are a function $G(\cdot)$ of the visual features. What would be a good choice for $G$? Recall that our aim is to recover an embedding for words that respects similarities in visual instantiations of words (\reffig{fig:teaser}). To capture this visual similarity, we model $G: v\rightarrow \{1,\cdots,N_K\}$ as a grouping function\footnote{Alternatively, one could regress directly to the feature values $v$. However, we found that the regression objective hurts performance.}. In practice, this function is learnt offline using clustering with K-means. That is, the outputs from clustering are the surrogate class labels used in \vwv{} training. Since we want our embeddings to reason about fine-grained visual grounding (\eg ``stares at'' and ``eats''), we cluster in the abstract scenes feature space (\refsec{sec:applications}). See \reffig{fig:clustering} for an illustration of what clustering captures. The parameter $N_K$ in K-means modulates the granularity at which we reason about visual grounding. 

\compactparagraph{Initialization:} We initialize the projection matrix parameters $W_{I}$ with those from training \wv{} on large text corpora. The hidden-to-output layer parameters are initialized randomly. Using \wv{} is advantageous for us in two ways: i) \wv{} embeddings have been shown to capture rich semantics and generalize to a large number of tasks in text. Thus, they provide an excellent starting point to finetune the embeddings to account for visual similarity as well. ii) Training on a large corpus gives us good coverage in terms of the vocabulary. \satwik{Further, since the gradients during backpropagation only affect parameters/embeddings for words seen during training, one can view \vwv{} as augmenting \wv{} with visual information when available. In other words, we retain the rich amount of non-visual information already present in it}\footnote{We verified empirically that this does not cause \emph{calibration} issues. Specifically, given a pair of words where one word was refined using visual information but the other was not (unseen during training), using \vwv{} for the former and \wv{} for the latter when computing similarities between the two outperforms using \wv{} for both.}. Indeed, we find that the random initialization does not perform as well as initialization with \wv{} when training \vwv.
 
\compactparagraph{Design Choices:}
Our model (\refsec{subsec:model}) admits choices of $w$ in a variety of forms such as full sentences or tuples of the form (Primary Object, Relation, Secondary Object). The exact choice of $w$ is made depending upon on what is natural for the task of interest. For instance, for common sense assertion classification and text-based image retrieval, $w$ is a phrase from a tuple, while for visual paraphrasing $w$ is a sentence. Given $w$, the choice of $S_w$ is also a design parameter tweaked depending upon the task. It could include all of $w$ (\eg., when learning from a phrase in the tuple) or a subset of the words (\eg., when learning from an $n$-gram context-window in a sentence). While the model itself is task agnostic, and only needs access to the words and visual context during training, the validation and test performances are calculated using the \vwv{} embeddings on a specific task of interest (\refsec{sec:experiments}). This is used to choose the hyperparameters $N_K$ and $N_H$.

\compactparagraph{Connections to \wv:} Our model can be seen as a multi-modal extension of the continuous bag of words (CBOW) \wv{} models. The CBOW \wv{} objective maximizes the likelihood $P(w|S_{w}, W_{I}, W_{O})$ for a word $w$ and its context $S_{w}$. On the other hand, we maximize the likelihood of the visual context given a set of words $S_{w}$ (\refeq{eqn:opt}). 


\section{Applications}\label{sec:applications}
We compare \vwv{} and \wv{} on the tasks of common sense assertion classification (\refsec{sec:cs_task}), visual paraphrasing (\refsec{sec:vp_data}), and text-based image retrieval (\refsec{sec:ret_task}). We give details of each task and the associated datasets below.
%
%

\subsection{Common Sense Assertion Classification}\label{sec:cs_task}
We study the relevance of \vwv{} to the common sense (CS) assertion classification task introduced by Vedantam \etal~\cite{vedantamLICCV15}. Given common sense tuples of the form (primary object or $t_P$, relation or $t_R$, secondary object or $t_S$)~\eg~(\texttt{boy}, \texttt{eats}, \texttt{cake}), the task is to classify it as plausible or not. The CS dataset contains $14,332$ TEST assertions (spanning $203$ relations) out of which $37\%$ are plausible, as indicated by human annotations. These TEST assertions are extracted from the MS COCO dataset~\cite{LinECCV14coco}, which contains real images and captions. Evaluating on this dataset allows us to demonstrate that visual grounding learnt from the abstract world generalizes to the real world.~\cite{vedantamLICCV15} approaches the task by constructing a multi-modal similarity function between TEST assertions whose plausibility is to be evaluated, and TRAIN assertions that are known to be plausible. The TRAIN dataset also contains $4260$ abstract scenes made from clipart depicting $213$ relations between various objects (20 scenes per relation). Each scene is annotated with one tuple that names the primary object, relation, and secondary object depicted in the scene. Abstract scene features (from~\cite{vedantamLICCV15}) describing the interaction between objects such as relative location, pose, absolute location, \etc are used for learning \vwv. More details of the features can be found in the appendix. We use the VAL set from~\cite{vedantamLICCV15} ($14,548$ assertions) to pick the hyperparameters. Since the dataset contains tuples of the form ($t_P$, $t_R$, $t_S$), we explore learning \vwv{} with \textbf{separate} models for each, and a \textbf{shared} model irrespective of the word being $t_P$, $t_R$, or $t_S$.

\begin{figure}
\centering
\includegraphics[width=\columnwidth]{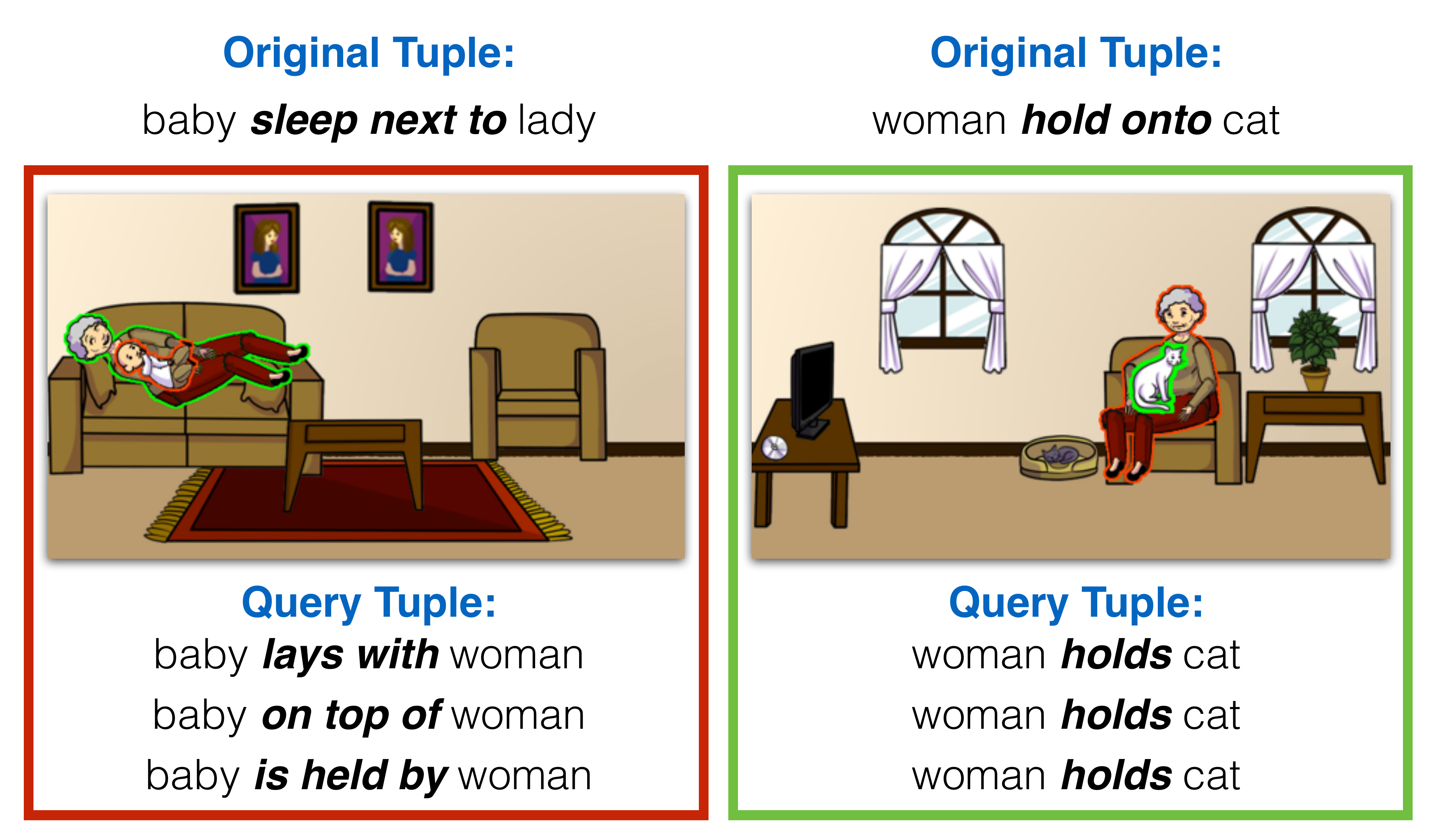}
\caption{Examples tuples collected for the text-based image retrieval task. Notice that multiple relations can have the same visual instantiation (left).}
\label{fig:spec}
\vspace{-15pt}
\end{figure}

\subsection{Visual Paraphrasing}\label{sec:vp_data}
\rama{Visual paraphrasing (VP), introduced by Lin and Parikh~\cite{Lin_2015_CVPR} is the task of determining if a pair of descriptions describes the same scene or two different scenes.} The dataset introduced by~\cite{Lin_2015_CVPR} contains $30,600$ pairs of descriptions, of which a third are positive (describe the same scene) and the rest are negatives. The TRAIN dataset contains $24,000$ VP pairs whereas the TEST dataset contains $6,060$ VP pairs. Each description contains three sentences. We use scenes and descriptions from Zitnick~\etal~\cite{Zitnick_2013_ICCV} to train \vwv{} models, similar to Lin and Parikh. The abstract scene feature set from~\cite{Zitnick_2013_ICCV} captures occurrence of objects, person attributes (expression, gaze, and pose), absolute spatial location and co-occurrence of objects, relative spatial location between pairs of objects, and depth ordering (3 discrete depths), relative depth and flip. We withhold a set of $1000$ pairs (333 positive and 667 negative) from TRAIN to form a VAL set to pick hyperparameters. Thus, our VP TRAIN set has $23,000$ pairs.

\subsection{Text-based Image Retrieval}\label{sec:ret_task}
In order to verify if our model has learnt the visual grounding of concepts, we study the task of text-based image retrieval. Given a query tuple, the task is to retrieve the image of interest by matching the query and ground truth tuples describing the images using word embeddings. For this task, we study the generalization of \vwv{} embeddings learnt for the common sense (CS) task,~\ie, there is no training involved. We augment the common sense (CS) dataset~\cite{vedantamLICCV15} (\refsec{sec:cs_task}) to collect three query tuples for each of the original 4260 CS TRAIN scenes. Each scene in the CS TRAIN dataset has annotations for which objects in the scene are the primary and secondary objects in the ground truth tuples. We highlight the primary and secondary objects in the scene and ask workers on AMT to name the primary, secondary objects, and the relation depicted by the interaction between them. Some examples can be seen in ~\reffig{fig:spec}. Interestingly, some scenes elicit diverse tuples whereas others tend to be more constrained. This is related to the notion of Image Specificity~\cite{jas2015specificity}. Note that the workers do not see the original (ground truth) tuple written for the scene from the CS TRAIN dataset. More details of the interface are provided in the appendix. We use the collected tuples as queries for performing the retrieval task. Note that the queries used at test time were never used for training \vwv.

\section{Experimental Setup}\label{sec:experiments}
%
%
%
We now explain our experimental setup. We first explain how we use our \vwv~or baseline \wv{} (word2vec)~model for the three tasks described above: common sense (CS), visual paraphrasing (VP), and text-based image retrieval. We also provide evaluation details. We then list the baselines we compare to for each task and discuss some design choices.
For all the tasks, we preprocess raw text by tokenizing using the NLTK toolkit~\cite{NLTK}. We implement \vwv{} as an extension of the Google C implementation of word2vec\footnote{https://code.google.com/p/word2vec/}.

\subsection{Common Sense Assertion Classification}\label{sec:exp_cs}
The task in common sense assertion classification (\refsec{sec:cs_task}) is to compute the plausibility of a test assertion based on its similarity to a set of tuples ($\Omega=\{t^i\}_{i=1}^{I}$) known to be plausible. Given a tuple $t'=$\texttt{(Primary Object $t'_P$, Relation $t'_R$, Secondary Object $t'_S$)} and a training instance $t^i$, the plausibility scores are computed as follows:
\begin{multline}
h(t', t^i) = W_{P}(t'_P)^T W_{P}(t^i_P) \\
+ W_{R}(t'_R)^T W_{R}(t^i_R) + W_{S}(t'_S)^T W_{S}(t^i_S) 
\end{multline}
where \satwik{$W_P, W_R, W_S$ represent the corresponding word embedding spaces}. The final text score is given as follows:
\begin{equation}
f(t') = \frac{1}{|I|} \sum_{i \in I} \max( h(t', t^i) - \delta, 0)
\end{equation}
where $i$ sums over the entire set of training tuples. We use the value of $\delta$ used by~\cite{vedantamLICCV15} for our experiments. 

~\cite{vedantamLICCV15} share embedding parameters across $t_P$, $t_R$, $t_S$ in their text based model. That is, $W_{P} = W_{R} = W_{S}$. We call this the \textbf{shared} model. When $W_P, W_R, W_S$ are learnt independently for ($t_P$, $t_R$, $t_S$), we call it the \textbf{separate} model.

The approach in~\cite{vedantamLICCV15} also has a visual similarity function that combines text and abstract scenes that is used along with this text-based similarity. We use the text-based approach for evaluating both \vwv{} and baseline \wv{}. However, we also report results including the visual similarity function along with text similarity from \vwv{}. In line with~\cite{vedantamLICCV15}, \satwik{we also evaluate our results using average precision (AP) as a performance metric.}

\subsection{Visual Paraphrasing}\label{sec:exp_vp}
In the visual paraphrasing task (\refsec{sec:vp_data}), we are given a pair of descriptions at test time. We need to assign a score to each pair indicating how likely they are to be paraphrases, \ie, describing the same scene. Following~\cite{Lin_2015_CVPR} we average word embeddings (\vwv{} or \wv{}) for the sentences and plug them into their text-based scoring function. This scoring function combines term frequency, word co-occurrence statistics and averaged word embeddings to assess the final paraphrasing score. The results are evaluated using average precision (AP) as the metric. \rama{While training both \vwv{} and \wv{} for the task, we append the sentences from the train set of ~\cite{Lin_2015_CVPR} to the original word embedding training corpus to handle vocabulary overlap issues.}

\subsection{Text-based Image Retrieval}\label{sec:exp_ir}
We compare \wv{} and \vwv{} on the task of text-based image retrieval (\refsec{sec:ret_task}). The task involves retrieving the target image from an image database, for a query tuple. Each image in the database has an associated ground truth tuple describing it. We use these to rank images by computing similarity with the query tuple. Given tuples of the form ($t_P$, $t_R$, $t_S$), we average the vector embeddings for all words in $t_P$, $t_R$, $t_S$. We then explore \textbf{separate} and \textbf{shared} models just as we did for common sense assertion classification. In the \textbf{separate} model, we first compute the cosine similarity between the query and the ground truth for $t_P$, $t_R$, $t_S$ separately and average the three similarities. In the \textbf{shared} model, we average the word embeddings for $t_P$, $t_R$, $t_S$ for query and ground truth and then compute the cosine similarity between the averaged embeddings. The similarity scores are then used to rank the images in the database for the query. We use standard metrics for retrieval tasks to evaluate: \texttt{Recall@1} (\texttt{R@1}), \texttt{Recall@5} (\texttt{R@5}), \texttt{Recall@10} (\texttt{R@10}) and median rank (\texttt{med R}) of target image in the returned result. 

\subsection{Baselines}
We describe some baselines in this subsection. In general, we consider two kinds of \wv{} models: those learnt from generic text, \eg., Wikipedia (\wv\wiki{}) and those learnt from visual text, \eg., MS COCO (\wv\coco),~\ie, text describing images. Embeddings learnt from visual text typically contain more visual information~\cite{vedantamLICCV15}. \vwv\wiki{} are \vwv{} embeddings learnt using \wv\wiki{} as an initialization to the projection matrix, while \vwv\coco{} are the \vwv{} embeddings learnt using \wv\coco{} as the initialization. In all settings, we are interested in studying the performance gains on using \vwv{} over \wv{}. Although our training procedure itself is task agnostic, we train separately on the common sense (CS) and the visual paraphrasing (VP) datasets. We study generalization of the embeddings learnt for the CS task on the text-based image retrieval task. Additional design choices pertaining to each task are discussed in~\refsec{sec:approach}.

\section{Results}\label{sec:results}
We present results on common sense (CS), visual paraphrasing (VP), and text-based image retrieval tasks. We compare our approach to various baselines as explained in \refsec{sec:experiments} for each application. Finally, we train our model using real images instead of abstract scenes, and analyze differences. More details on the effect of hyperparameters on performance (for CS and VP) can be found in the appendix.
\begin{table} \footnotesize
\setlength{\tabcolsep}{7.5pt}
\begin{center}
\begin{tabular}{@{} l  c  c @{}}
\toprule
Approach & common sense AP (\%)\\
\midrule
\vwv\wiki{} (shared) & 72.2\\
\vwv\wiki{} (separate) & 74.2\\
\satwik{\vwv\coco{} (shared) + vision} & \satwik{74.2}\\
\satwik{\vwv\coco{} (shared)} & \satwik{74.5}\\ 
\vwv\coco{} (separate) & \textbf{74.8}\\ 
\vwv\coco{} (separate) + vision & \textbf{75.2}\\
\wv\wiki{} (from ~\cite{vedantamLICCV15}) &   68.4\\
\wv\coco{} (from ~\cite{vedantamLICCV15})  &   72.2\\
\wv\coco{} + vision (from ~\cite{vedantamLICCV15}) & 73.6\\
\bottomrule
\end{tabular}
\caption{Performance on the common sense task of ~\cite{vedantamLICCV15}}
\label{table:cs}
\vspace{-20pt}
\end{center}
\end{table}

\subsection{Common Sense Assertion Classification}\label{subsec:cs_res}
\rama{We first present our results on the common sense assertion classification task (\refsec{sec:cs_task}). We report numbers with a fixed hidden layer size, $N_H=200$ (to be comparable to~\cite{vedantamLICCV15}) in Table.~\ref{table:cs}.} We use $N_K=25$, which gives the best performance on validation. \satwik{We handle tuple elements, $t_P$, $t_R$ or $t_S$, with more than one word by placing each word in a separate window (\ie $|S_{w}|=1$). For instance, the element ``lay next to'' is trained by predicting the associated visual context thrice with ``lay'', ``next'' and ``to'' as inputs.} Overall, we find an increase of 2.6\% with \vwv\coco{} (separate) model over the \wv\coco{} model used in ~\cite{vedantamLICCV15}. We achieve larger gains (5.8\%) with \vwv\wiki{} over \wv\wiki{}. \rama{Interestingly, the tuples in the common sense task are extracted from the MS COCO~\cite{LinECCV14coco} dataset. Thus, this is an instance where \vwv{} (learnt from abstract scenes) generalizes to text describing real images.}

\rama{Our \vwv\coco{} (both shared and separate) embeddings outperform the joint \wv\coco~+~vision model from~\cite{vedantamLICCV15} that reasons about visual features for a given test tuple, which we do not}. Note that both models use the same training and validation data, which suggests that our \vwv{} model captures the grounding better than their multi-modal text~+~visual similarity model. Finally, we sweep for the best value of $N_H$ for the validation set and find that \vwv\coco{} (separate) gets the best AP of 75.4\% on TEST with $N_H=50$. \rama{This is our best performance on this task.}

\compactparagraph{Separate vs. Shared:} We next compare the performance when using the \textbf{separate} and \textbf{shared} \vwv{} models. We find that \vwv\coco{} (separate) does better than \vwv\coco{} (shared) (74.8\% vs. 74.5\%), presumably because the embeddings can specialize to the semantic roles words play when participating in $t_P$, $t_R$ or $t_S$. In terms of \textbf{shared} models alone, \vwv\coco{} (shared) achieves a gain in performance of 2.3\% over the \wv\coco{} model of~\cite{vedantamLICCV15}, whose textual models are all shared.

\begin{figure}[t]
\includegraphics[width=\columnwidth, trim={0 0.5cm 0 0}, clip]{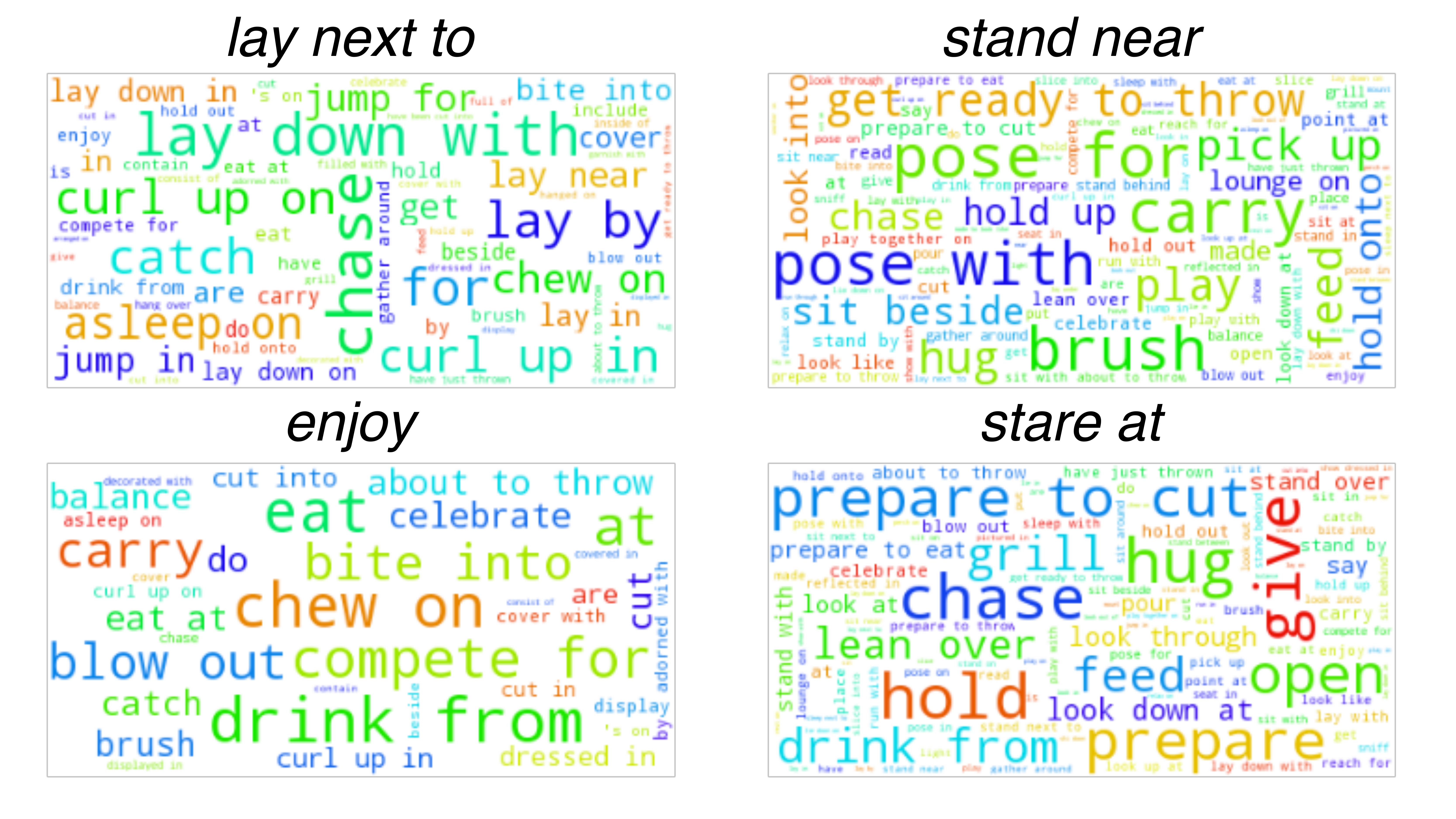}
\vspace{-20pt}
\caption{Visualization of the clustering used to supervise \vwv{} training. Relations that co-occur more often in the same cluster appear bigger than others. Observe how semantically close relations co-occur the most, \textit{e.g.}, \texttt{eat}, \texttt{drink}, \texttt{chew on} for the relation \texttt{enjoy}.}
\label{fig:clustering}
\vspace{-18pt}
\end{figure}

\compactparagraph{What Does Clustering Capture?} We next visualize the semantic relatedness captured by clustering in the abstract scenes feature space (\reffig{fig:clustering}). Recall that clustering gives us surrogate labels to train \vwv. For the visualization, we pick a relation and display other relations that co-occur the most with it in the same cluster. Interestingly, words like ``prepare to cut'', ``hold'', ``give'' occur often with ``stare at''. Thus, we discover the fact that when we ``prepare to cut'' something, we also tend to ``stare at'' it. Reasoning about such notions of semantic relatedness using purely textual cues would be prohibitively difficult. We provide more examples in the appendix.

\subsection{Visual Paraphrasing}
\begin{figure}[t]
\includegraphics[width=\columnwidth,page=2]{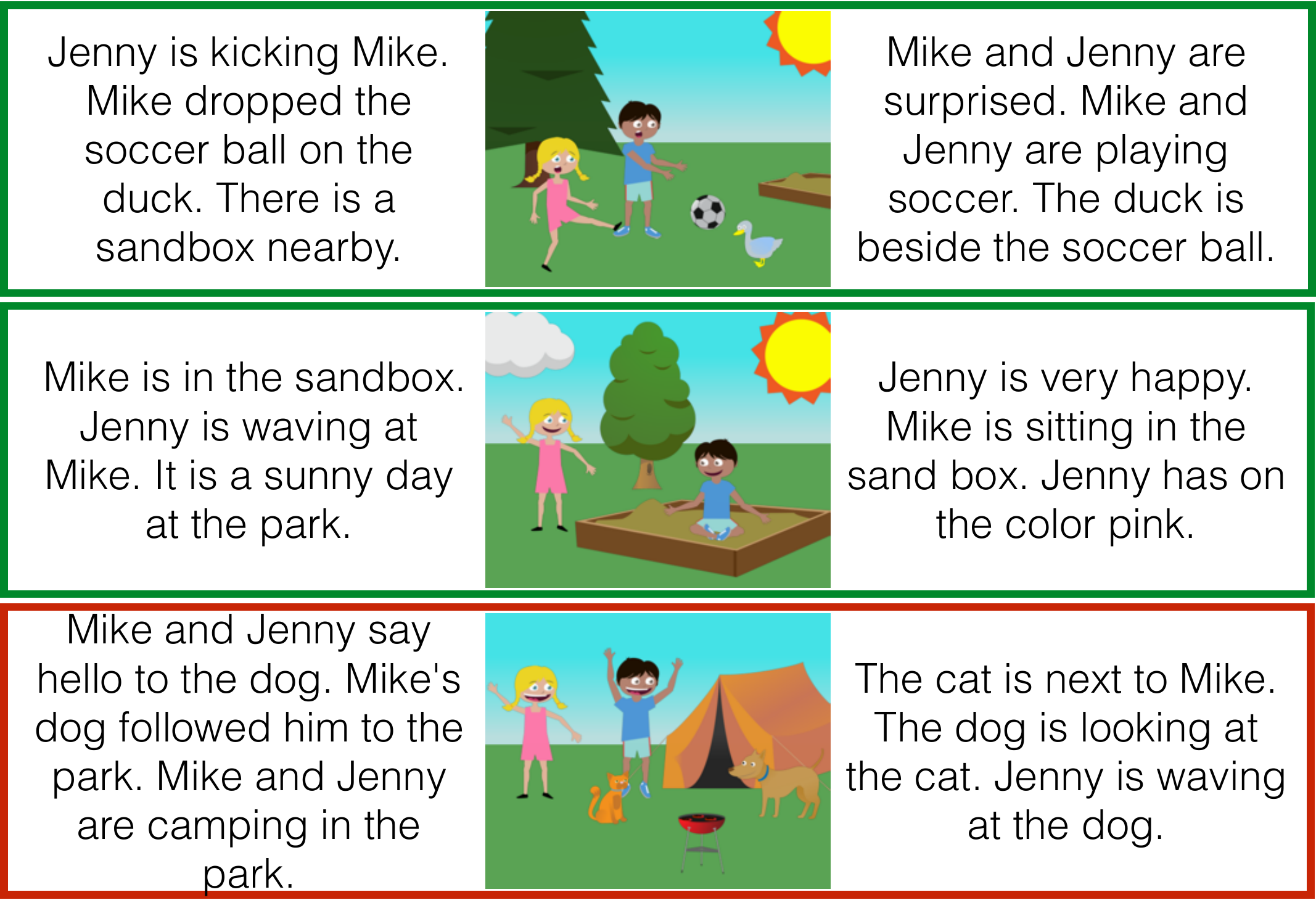}
\caption{\satwik{The visual paraphrasing task is to identify if two textual descriptions are paraphrases of each other. Shown above are three positive instances, \textit{i.e.}, the descriptions (left, right) actually talk about the same scene (center, shown for illustration, not avaliable as input). Green boxes show two cases where \vwv{} correctly predicts and \wv{} does not, while red box shows the case where both \vwv{} and \wv{} predict incorrectly. Note that the red instance is tough as the textual descriptions do not intuitively seem to be talking about the same scene, even for a human reader.}}
\label{fig:bar}
\end{figure}

We next describe our results on the Visual Paraphrasing (VP) task (\refsec{sec:vp_data}). The task is to determine if a pair of descriptions are describing the same scene. Each description has three sentences. Table.~\ref{table:vp} summarizes our results and compares performance to \wv{}. We vary the size of the context window $S_{w}$ and check performance on the VAL set. We obtain best results with the entire description as the context window $S_{w}$, $N_H=200$, and $N_K=100$. Our \vwv{} models give an improvement of 0.7\% on both \wv\wiki{} and \wv\coco{} respectively. In comparison to \wv\wiki{} approach from~\cite{Lin_2015_CVPR}, we get a larger gain of 1.2\% with our \vwv\coco{} embeddings\footnote{Our implementation of~\cite{Lin_2015_CVPR} performs 0.3\% higher than that reported in ~\cite{Lin_2015_CVPR}.}. Lin and Parikh~\cite{Lin_2015_CVPR} imagine the visual scene corresponding to text to solve the task. Their combined text~+~imagination model performs 0.2\% better (95.5\%) than our model. \satwik{Note that our approach does not have the additional expensive step of generating an imagined visual scene for each instance at test time.} Qualitative examples of success and failure cases are shown in \reffig{fig:bar}.

\compactparagraph{Window Size:} Since the VP task is on multi-sentence descriptions, it gives us an opportunity to study how size of the window ($S_{w}$) used in training affects performance. We evaluate the gains obtained by using window sizes of entire description, single sentence, 5 words, and single word respectively. We find that description level windows and sentence level windows give equal gains. However, performance tapers off as we reduce the context to 5 words (0.6\% gain) and a single word (0.1\% gain). This is intuitive, since VP requires us to reason about entire descriptions to determine paraphrases. Further, since the visual features in this dataset are scene level (and not about isolated interactions between objects), the signal in the hidden layer is stronger when an entire sentence is used.

\begin{table} \footnotesize
\setlength{\tabcolsep}{10pt}
\begin{center}
\begin{tabular}{@{} l  c  c @{}}
\toprule
Approach & Visual Paraphrasing AP (\%)\\
\midrule
\wv\wiki{} (from ~\cite{Lin_2015_CVPR}) & 94.1\\
\wv\wiki{}  &  94.4\\
\wv\coco{}  &   94.6\\
\vwv\wiki{} & 95.1\\ 
\vwv\coco{} & \textbf{95.3}\\
\bottomrule
\end{tabular}
\caption{Performance on visual paraphrasing task of~\cite{Lin_2015_CVPR}.}
\label{table:vp}
\vspace{-20pt}
\end{center}
\end{table}

\subsection{Text-based Image Retrieval}
We next present results on the text-based image retrieval task (\refsec{sec:ret_task}). This task requires visual grounding as the query and the ground truth tuple can often be different by textual similarity, but could refer to the same scene (\reffig{fig:spec}). As explained in \refsec{sec:ret_task}, we study generalization of the embeddings learnt during the commonsense experiments to this task. Table.~\ref{table:ir} presents our results. Note that \vwv{} here refers to the embeddings learnt using the CS dataset. We find that the best performing models are \vwv\wiki{} (shared) (as per \texttt{R@1}, \texttt{R@5}, \texttt{medR}) and \vwv\coco{} (separate) (as per \texttt{R@10}, \texttt{medR}). These get \texttt{Recall@10} scores of $\approx$49.5\% whereas the baseline \wv\wiki{} and \wv\coco{} embeddings give scores of 45.4\% and 47.6\%, respectively. 
\begin{table} \footnotesize
\setlength{\tabcolsep}{5pt}
\begin{center}
\begin{tabular}{@{} l  c  c  c  c @{}}
\toprule
Approach & \texttt{R@1} (\%) & \texttt{R@5} (\%) & \texttt{R@10} (\%) & \texttt{med R}\\
\midrule
\wv\wiki{}  &  14.6 & 34.4 & 45.4 & 13\\
\wv\coco{}  &  15.3 & 35.2 & 47.6 & 11\\
\vwv\wiki{} (shared) & 15.5 & 37.2 & 49.3 & \textbf{10}\\ 
\vwv\coco{} (shared) & \textbf{15.7} & \textbf{37.7} & 47.6 & \textbf{10}\\
\vwv\wiki{} (separate) & 14.0 & 32.7 & 43.5 & 15\\ 
\vwv\coco{} (separate) & 15.4 & 37.6 & \textbf{49.5} & \textbf{10}\\
\bottomrule
\end{tabular}
\caption{Performance on text-based image retrieval. \texttt{R@x}: \textbf{higher} is better, \texttt{medR}: \textbf{lower} is better}
\label{table:ir}
\vspace{-25pt}
\end{center}
\end{table}

\subsection{Real Image Experiment} \label{sub:real-im}
Finally, we test our \vwv{} approach with real images on the CS task, to evaluate the need to learn fine-grained visual grounding via abstract scenes. Thus, instead of semantic features from abstract scenes, we obtain surrogate labels by clustering real images from the MS COCO dataset using \texttt{fc7} features from the VGG-16~\cite{DBLP:journals/corr/SimonyanZ14a} CNN. We cross validate to find the best number of clusters and hidden units. We perform real image experiments in two settings: 1) We use all of the MS COCO dataset after removing the images whose tuples are in the CS TEST set of ~\cite{vedantamLICCV15}. This gives us a collection of $\approx 76$K images to learn \vwv{}. MS COCO dataset has a collection of $5$ captions for each image. We use all these five captions with sentence level context\footnote{We experimented with other choices but found this works best.} windows to learn \vwv\texttt{80K}. \satwik{2) We create a real image dataset by collecting 20 real images from MS COCO and their corresponding tuples, randomly selected for each of $213$ relations from the VAL set (\refsec{sec:exp_cs}). Analogous to the CS TRAIN set containing abstract scenes, this gives us a dataset of 4260 real images along with an associate tuple, depicting the 213 CS VAL relations. We refer to this model as \vwv\texttt{4K}.}

\satwik{We report the gains in performance over \wv{} baselines in both scenario 1) and 2) for the common sense task. We find that using real images gives a best-case performance of 73.7\% starting from \wv\coco{} for \vwv\texttt{80K} (as compared to 74.8\% using CS TRAIN abstract scenes). For \vwv\texttt{4K}\coco, the performance on the validation actually goes down during training. If we train \vwv\texttt{4K} starting with generic text based \wv\wiki{}, we get a performance of 70.8\% (as compared to 74.2\% using CS TRAIN abstract scenes). This shows that abstract scenes are better at visual grounding as compared to real images, due to their rich semantic features.}

\section{Discussion}\label{sec:discussion}
Antol \etal~\cite{Antol_2014} have studied generalization of classification models learnt on abstract scenes to real images. The idea is to transfer fine-grained concepts that are easier to learn in the fully-annotated abstract domain to tasks in the real domain. Our work can also be seen as a method of studying generalization. One can view \vwv{} as a way to transfer knowledge learnt in the abstract domain to the real domain, via text embeddings (which are shared across the abstract and real domains). Our results on commonsense assertion classification show encouraging preliminary evidence of this.

We next discuss some considerations in the design of the model. A possible design choice when learning embeddings could have been to construct a triplet loss function, where the similarity between a tuple and a pair of visual instances can be specified. That is, given a textual instance A, and two images B and C (where A describes B, and not C), one could construct a loss that enforces $sim(A, B) > sim(A, C)$, and learn joint embeddings for words and images. However, since we want to learn hidden semantic relatedness (\eg.``eats", ``stares at"), there is no explicit supervision available at train time on which images and words should be related. Although the visual scenes and associated text inherently provide information about related words, they do not capture the unrelatedness between words, \ie, we do not have negatives to help us learn the semantics. 

We can also understand \vwv{} in terms of data augmentation. With infinite text data describing scenes, distributional statistics captured by \wv{} would reflect all possible visual patterns as well. In this sense, there is nothing special about the visual grounding. The additional modality helps to learn complimentary concepts while making efficient use of data. Thus, the visual grounding can be seen as augmenting the amount of textual data. 

\section{Conclusion}\label{sec:conclusion}
We learn visually grounded word embeddings (\vwv{}) from abstract scenes and associated text. Abstract scenes, being trivially fully annotated, give us access to a rich semantic feature space. We leverage this to uncover visually grounded notions of semantic relatedness between words that would be difficult to capture using text alone or using real images. We demonstrate the visual grounding captured by our embeddings on three applications that are in text, but benefit from visual cues: 1) common sense assertion classification, 2) visual paraphrasing, and 3) text-based image retrieval. Our method outperforms word2vec (\wv{}) baselines on all three tasks. Further, our method can be viewed as a modality to transfer knowledge from the abstract scenes domain to the real domain via text. Our datasets, code, and \vwv{} embeddings are available for public use.

\compactparagraph{Acknowledgments: }
This work was supported in part by the The Paul G. Allen Family Foundation via an award to D.P., ICTAS at Virginia Tech via an award to D.P., a Google Faculty Research Award to D.P. the Army Research Office YIP Award to D.P, and ONR grant N000141210903.

\appendix
\section*{Appendix}
We present detailed performance results of Visual Word2Vec (\vwv{}) on all three tasks : 
\begin{itemize}
\item Common sense assertion classification (\refsec{sec:commonsense})
\item Visual paraphrasing (\refsec{sec:paraphrase})
\item Text-based image retrieval (\refsec{sec:retrieval})
\end{itemize}
Specifically, we study the affect of various hyperparameters like number of surrogate labels ($K$), number of hidden layer nodes ($N_H$), \etc., on the performance of both \vwv\coco{} and \vwv\wiki{}. 
We remind the reader that \vwv\coco{} models are initialized with \wv{} learnt on \emph{visual text}, \ie., MSCOCO captions in our case while \vwv\wiki{} models are initialized with \wv{} learnt on generic Wikipedia text. 
We also show few visualizations and examples to qualitatively illustrate why \vwv{} performs better in these tasks that are ostentatiously in text, but benefit from visual cues.
We conclude by presenting the results of training on real images (\refsec{sec:real-images}). We also show a comparison to the model from Ren~\etal, who also learn word2vec with visual grounding.

\section{Common Sense Assertion Classification}
\label{sec:commonsense}
Recall that the common sense assertion classification task~\cite{vedantamLICCV15} is to determine if a tuple of the form (primary object or P, relation or R, secondary object or S) is plausible or not. 
In this section, we first describe the abstract visual features used by~\cite{vedantamLICCV15}.
We follow it with results for \vwv\coco{}, both \textbf{shared} and \textbf{separate} models, by varying the number of surrogate classes $K$. 
We next discuss the effect of number of hidden units $N_H$ which can be seen as the complexity of the model.
We then vary the amount of training data and study performance of \vwv\coco. 
Learning separate word embeddings for each of these specific roles, \ie., P, R or S results in \textbf{separate} models while learning single embeddings for all of them together gives us \textbf{shared} models. 
Additionally, we also perform and report similar studies for \vwv\wiki{}.
Finally, we visualize the clusters learnt for the common sense task through word clouds, similar to Fig. 4 in the main paper. 
\subsection{Abstract Visual Features}
We describe the features extracted from abstract scenes for the task of common sense assertion classification. Our visual features are essentially the same as those used by ~\cite{vedantamLICCV15}:
a) Features corresponding to primary and secondary object, \ie., P and S respectively. 
These include type (category ID and instance ID), absolute location modeled via Gaussian Mixture Model (GMM), orientation, attributes and poses for both P and S present in the scene.
We use Gaussian Mixture at hands and foot locations to model pose, measuring relative positions and joint locations. Human attributes are age (5 discrete values), skin color (3 discrete values) and gender (2 discrete values). Animals have 5 discrete poses. Human pose features are constructed
using keypoint locations. 
b) Features corresponding to relative location of P and S, once again modeled using Gaussian Mixture Models. These features are normalized by the flip and depth of the primary object, which results in the features being asymmetric. We compute these with respect to both P and S to make the features symmetric.
c) Features related to the presence of other objects in the scene, \ie., category ID and instance ID for all the other objects.
Overall the feature vector is of dimension $1222$.
\subsection{Varying number of clusters $K$}
\label{subsec:vary-clusters}
\compactparagraph{Intuition: }
We cluster the images in the semantic clipart feature space to get surrogate labels. We use these labels as visual context, and predict them using words to enforce visual grounding. Hence, we study the influence of the number of surrogate classes relative to the number of images. This is indicative of how coarse/detailed the visual grounding for a task needs to be.
\begin{figure*}[ht]
    \centering
    \begin{subfigure}[b]{\columnwidth}
        \includegraphics[width=\columnwidth, trim={1cm 0cm 0cm 0cm}]{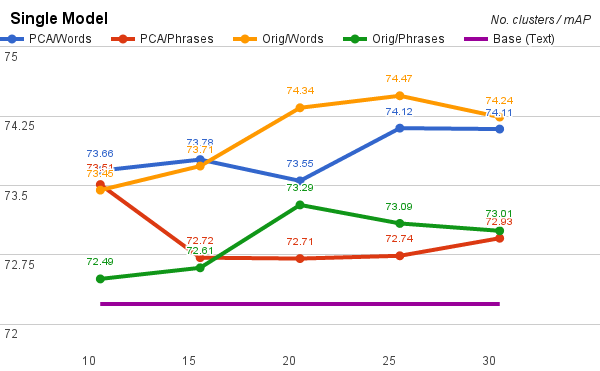}
        \caption{Shared model}
        \label{fig:shared-model-coco}
    \end{subfigure}
	\begin{subfigure}[b]{\columnwidth}
        \includegraphics[width=\columnwidth, trim={0cm 0cm 1cm 0cm}]{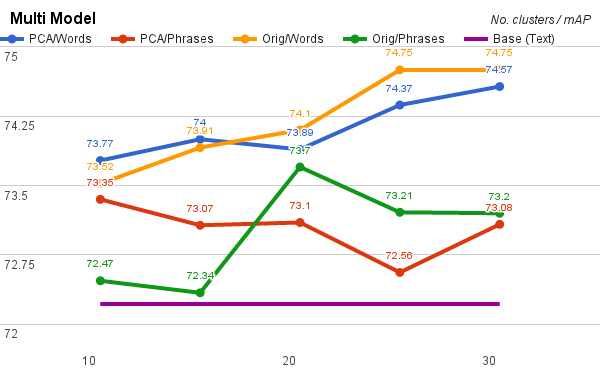}
        \caption{Separate model}
        \label{fig:separate-model-coco}
    \end{subfigure}    
    \caption{Common sense task performance for \textbf{shared} and \textbf{separate} models on varying the number of surrogate classes. $K$ determines the detail in visual information used to provide visual grounding. Note that the performance increases and then either saturates or decreases. Low $K$ results in an uninformative/noisy visual context while high $K$ results in clusters with insufficient grounding. Also note that \textbf{separate} models outperform the \textbf{shared} models. This indicates that \vwv{} learns different semantics specific to the role each word plays, \ie. P, R or S.\label{fig:vary-clusters-coco}}
\end{figure*}

\compactparagraph{Setup:}
We train \vwv{} models by clustering visual features with and without dimensionality reduction through Principal Component Analysis (PCA), giving us \texttt{Orig} and \texttt{PCA} settings, respectively.
Notice that each of the elements of tuples, \ie., P, R or S could have multiple words, \eg., \emph{lay next to}.
We handle these in two ways:
a) Place each of the words in separate windows and predict the visual context repeatedly. Here, we train by predicting the same visual context for \emph{lay}, \emph{next}, \emph{to} thrice. This gives us the \texttt{Words} setting.
b) Place all the words in a single window and predict the visual context for the entire element only once. This gives the \texttt{Phrases} setting.
We explore the cross product space of settings a) and b). \texttt{PCA/Phrases} (red in \reffig{fig:vary-clusters-coco}) refers to the model trained by clustering the dimensionality reduced visual features and handling multi-word elements by including them in a single window.
We vary the number of surrogate classes from $15$ to $35$ in steps of $5$, re-train \vwv{} for each $K$, and report the accuracy on the common sense task. The number of hidden units $N_H$ is kept fixed to $200$ to be comparable to the text-only baseline reported in~\cite{vedantamLICCV15}.
\reffig{fig:vary-clusters-coco} shows the performance on the common sense task as $K$ varies for both \textbf{shared} and \textbf{separate} models in four possible configurations each, as described above.

\compactparagraph{Observations:}
\begin{itemize}[leftmargin=*, itemsep=0pt]
\item As $K$ varies, the performance for both \textbf{shared} and \textbf{separate} models increases initially and then either saturates or decreases. For a given dataset, low values of $K$ result in the visual context being too coarse to learn the visual grounding. On the other hand, $K$ being too high results in clusters which do not capture visual semantic relatedness. We found the best model to have around $25$ clusters in both the cases.

\item \texttt{Words} models perform better than \texttt{Phrases} models in both cases. 
Common sense task involves reasoning about the specific role (P, R or S) each word plays.
For example, \texttt{(man, eats, sandwich)} is plausible while \texttt{(sandwich, eats, sandwich)} or \texttt{(man, sandwich, eats)} is not.
Potentially, \vwv{} could learn these roles in addition to the learning semantic relatedness between the words.
This explains why \textbf{separate} models perform better than \textbf{shared} models, and \texttt{Words} outperform \texttt{Phrases} setting.

\item For lower $K$, \texttt{PCA} models dominate over \texttt{Orig} models while the latter outperforms as $K$ increases.
As low values of $K$ correspond to coarse visual information, surrogate classes in \texttt{PCA} models could be of better quality and thus help in learning the visual semantics.
\end{itemize}
\begin{figure*}[ht]
    \centering
	\begin{subfigure}[b]{\columnwidth}
        \includegraphics[width=1.05\columnwidth, trim={0cm 0cm 2pt 0cm}, clip]{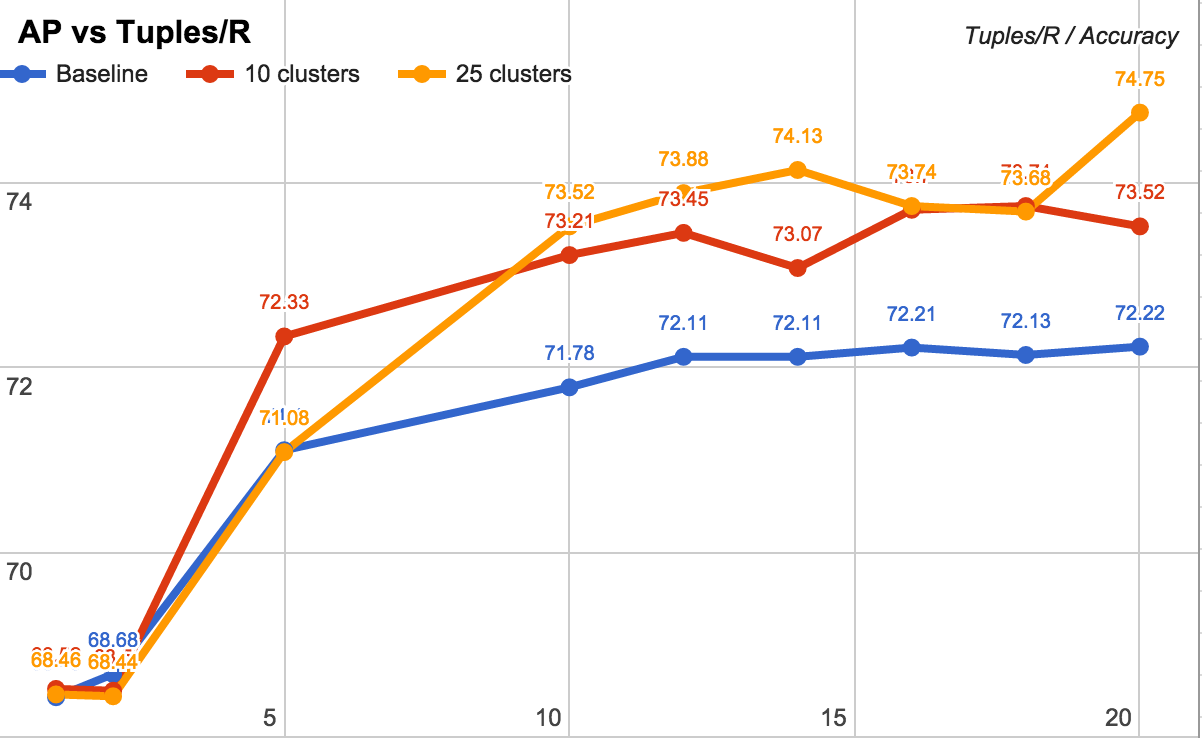}
        \caption{Varying the number of abstract scenes per relation, $n_T$}
        \label{fig:vary-data-perR}
    \end{subfigure}\quad
    \begin{subfigure}[b]{\columnwidth}
        \includegraphics[width=1.05\columnwidth, trim={0cm 0cm 1cm 0cm}, clip]{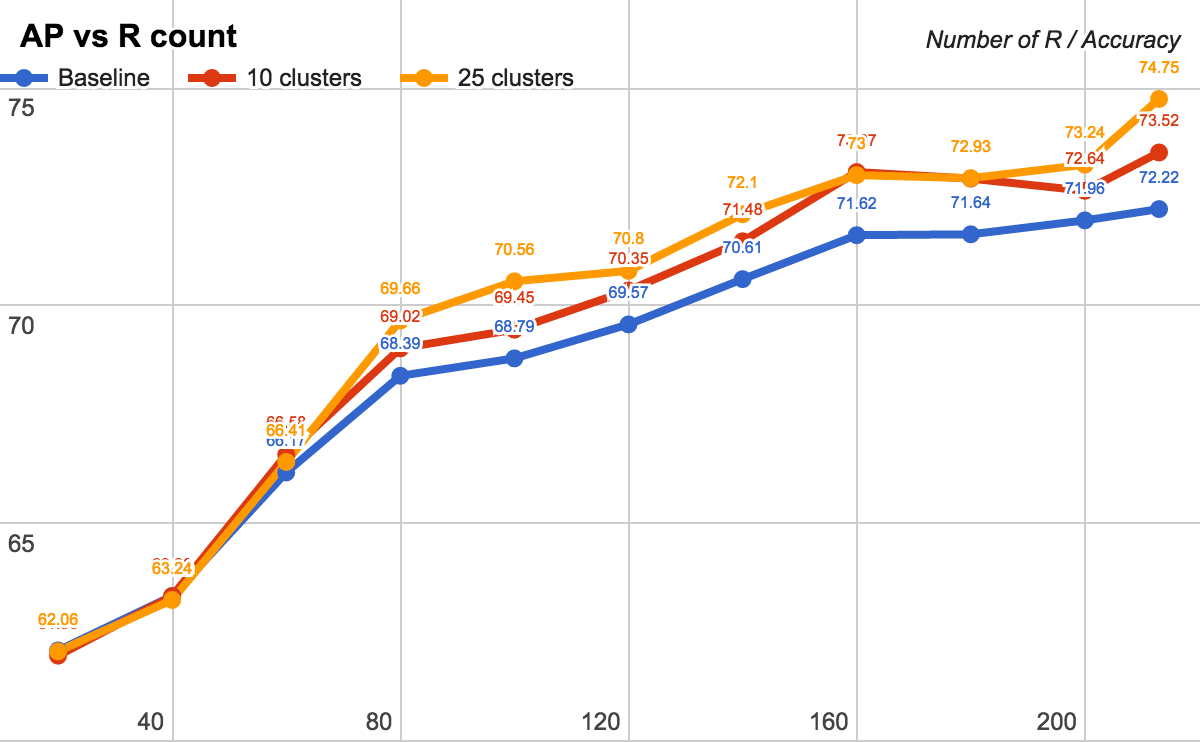}
        \caption{Varying the number of relations, $n_R$}
        \label{fig:vary-data-R}
    \end{subfigure}
    \caption{Performance on common sense task, varying the size of training data. Note the performance saturating as $n_T$ increases (left) while it increases steadily with increasing $n_R$ (right). Learning visual semantics benefits from training on more relations over more examples per relation. In other words, \emph{breadth} of concepts is more crucial than the \emph{depth} for learning visual grounding through \vwv{}. As the \wv{} baseline exhibits similar behavior, we conclude the same for learning semantics through text.}
    \label{fig:vary-data}
\end{figure*}

\subsection{Varying number of hidden units $N_H$}
\compactparagraph{Intuition: }
One of the model parameters for our \vwv{} is the number of hidden units $N_H$. 
This can be seen as the capacity of the model. 
We vary $N_H$ while keeping the other factors constant during training to study its affect on performance of the \vwv{} model.

\begin{figure}[ht]
    \centering
    \includegraphics[width=\columnwidth, trim={2pt 0pt 5pt 2pt}, clip]{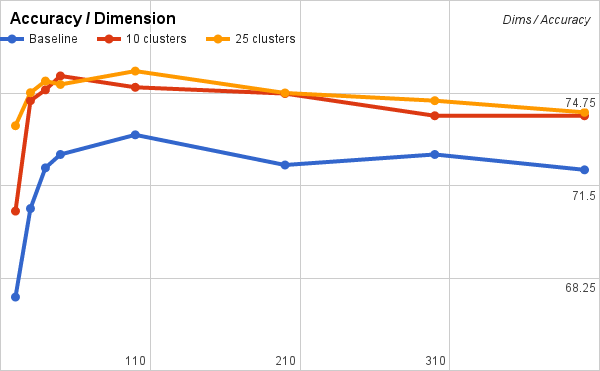}
    \caption{Performance on common sense task varying the number of hidden units $N_H$. This determines the complexity of the model used to learn visual semantics. Observe that models with low complexity perform the worst. Performance first rises reaching a peak and then decreases, for a fixed size of training data. Low end models do not capture visual semantics well while high end models overfit for the given data.}
    \label{fig:vary-dims-coco}
\end{figure}

\compactparagraph{Setup: }
To understand the role of $N_H$, we consider two \vwv{} models trained separately with $K$ set to $10$ and $25$ respectively.
Additionally, both of these are \textbf{separate} models with \texttt{Orig/Words} configuration (see \refsec{subsec:vary-clusters}).
We particularly choose these two settings as the former is trained with a very coarse visual semantic information while the latter is the best performing model.
Note that as~\cite{vedantamLICCV15} fix the number of hidden units to $200$ in their evaluation, we cannot directly compare the performance to their baseline.
We, therefore, recompute the baselines for each value of $N_H \in \{20, 30, 40, 50, 100, 200, 400\}$ and use it to compare our two models, as shown in \reffig{fig:vary-dims-coco}.

\compactparagraph{Observations: }
Models of low complexity, \ie., low values of $N_H$, perform the worst.
This could be due to the inherent limitation of low $N_H$ to capture the semantics, even for \wv{}.
On the other hand, high complexity models also perform poorly, although better than the low complexity models.
The number of parameters to be learnt, \ie. $W_I$ and $W_O$, increase linearly with $N_H$.
Therefore, for a finite amount of training data, models of high complexity tend to overfit resulting in drop in performance on an unseen test set.
The baseline \wv{} models also follow a similar trend.
It is interesting to note that the improvement of \vwv{} over \wv{} for less complex models (smaller $N_H$) is at $5.32\%$ (for $N_H=20$) as compared to $2.6\%$ (for $N_H=200$).
In other words, lower complexity models benefit more from the \vwv{} enforced visual grounding.
In fact, \vwv{} of low complexity $(N_H, K)=(20, 25)$, outperforms the best \wv{} baseline across all possible settings of model parameters.
This provides a strong evidence for the usefulness of visually grounding word embeddings in capturing visually-grounded semantics better.

\begin{table*}[t] \footnotesize
\setlength{\tabcolsep}{5.5pt}
\begin{center}
\begin{tabular}{@{} l  c c c c c c @{}}
\toprule
\multicolumn{7}{c}{\vwv\coco}\\
Model & $N_H$ & Baseline & \texttt{Descs} & \texttt{Sents} & \texttt{Winds} & \texttt{Words}\\
\midrule
\texttt{Orig} & \multirow{2}{*}{$50$} & \multirow{2}{*}{94.6} & \textbf{95.0} & \textbf{95.0} & 94.9 & 94.8\\
\texttt{PCA} &  &  & 94.9 & \textbf{95.1} & 94.7 & 94.8\\
\midrule
\texttt{Orig} & \multirow{2}{*}{$100$} & \multirow{2}{*}{94.6} & \textbf{95.3} & 95.1 & 95.1 & 94.9\\
\texttt{PCA} &  &  & \textbf{95.3} & \textbf{95.3} & 94.8 & 95.0\\
\midrule
\texttt{Orig} & \multirow{2}{*}{$200$} & \multirow{2}{*}{94.6} & 95.1 & \textbf{95.3} & 95.2 & 94.9\\
\texttt{PCA} &  &  & \textbf{95.3} & \textbf{95.3} & 95.2 & 94.8\\
\bottomrule
\end{tabular}
\begin{tabular}{@{} l  c c c c c c @{}}
\toprule
\multicolumn{7}{c}{\vwv\wiki}\\
Model & $N_H$ & Baseline & \texttt{Descs} & \texttt{Sents} & \texttt{Winds} & \texttt{Words}\\
\midrule
\texttt{Orig} & \multirow{2}{*}{$50$} & \multirow{2}{*}{94.2} & \textbf{94.9} & 94.8 & 94.7 & 94.7\\
\texttt{PCA} &  &  & \textbf{94.9} & \textbf{94.9} & 94.7 & 94.8\\
\midrule
\texttt{Orig} & \multirow{2}{*}{$100$} & \multirow{2}{*}{94.3} & \textbf{95.0} & 94.8 & 94.7 & 94.6\\
\texttt{PCA} &  &  & \textbf{95.1} & 94.9 & 94.7 & 94.7\\
\midrule
\texttt{Orig} & \multirow{2}{*}{$200$} & \multirow{2}{*}{94.4} & \textbf{95.1} & 94.8 & 94.7 & 94.5\\
\texttt{PCA} &  &  & \textbf{95.1} & 95.0 & 94.7 & 94.6\\
\bottomrule
\end{tabular}
\caption{Performance on the Visual Paraphrase task for \vwv\coco{} (left) and \vwv\wiki{} (right).}
\label{tab:vp}
\vspace{-25pt}
\end{center}
\end{table*}
\subsection{Varying size of training data}
\compactparagraph{Intuition: }
We next study how varying the size of the training data affects performance of the model. The idea is to analyze whether more data about relations would help the task, or more data \emph{per} relation would help the task. 

\compactparagraph{Setup: }
We remind the reader that \vwv{} for common sense task is trained on CS TRAIN dataset that contains 4260 abstract scenes made from clipart depicting 213 relations between various objects (20 scenes per relation). 
We identify two parameters: the number of relations $n_R$ and the number of abstract scenes per relation $n_T$.
Therefore, CS TRAIN dataset originally has $(n_T, n_R) = (20, 213)$.
We vary the training data size in two ways:
a) Fix $n_R=213$ and vary $n_T \in \{1, 2, 5, 10, 12, 14, 16, 18, 20\}$.
b) Fix $n_T=20$ and vary $n_R$ in steps of $20$ from $20$ to $213$.
These cases denote two specific situations--the former limits the model in terms of how much it knows about each relation, \ie. its \emph{depth}, keeping the number of relations, \ie. its \emph{breadth}, constant; while the latter limits the model in terms of how many relations it knows, \ie., it limits the \emph{breadth} keeping the \emph{depth} constant.
Throughout this study, we select the best performing \vwv{} model with $(K, N_H)=(25, 200)$ in the \texttt{Orig/Words} configuration.
\reffig{fig:vary-data-perR} shows the performance on the common sense task when $n_R$ is fixed while \reffig{fig:vary-data-R} is the performance when $n_T$ is fixed.

\compactparagraph{Observations: }
The performance increases with the increasing size of training data in both the situations when $n_T$ and $n_R$ is fixed. 
However, the performance saturates in the former case while it increases with almost a linear rate in the latter.
This shows that \emph{breadth} helps more than the \emph{depth} in learning visual semantics.
In other words, training with more relations and fewer scenes per relation is more beneficial than training with fewer relations and more scenes per relation.
To illustrate this, consider performance with approximately around half the size of the original CS TRAIN dataset.
In the former case, it corresponds to $73.5\%$ at $(n_T, n_R)=(10, 213)$ while $70.6\%$ at $(n_T, n_R)=(20, 100)$ in the latter.
Therefore, we conclude that the model learns semantics better with more concepts (relations) over more instances (abstract scenes) per concept.
\begin{figure*}
\includegraphics[width=\textwidth]{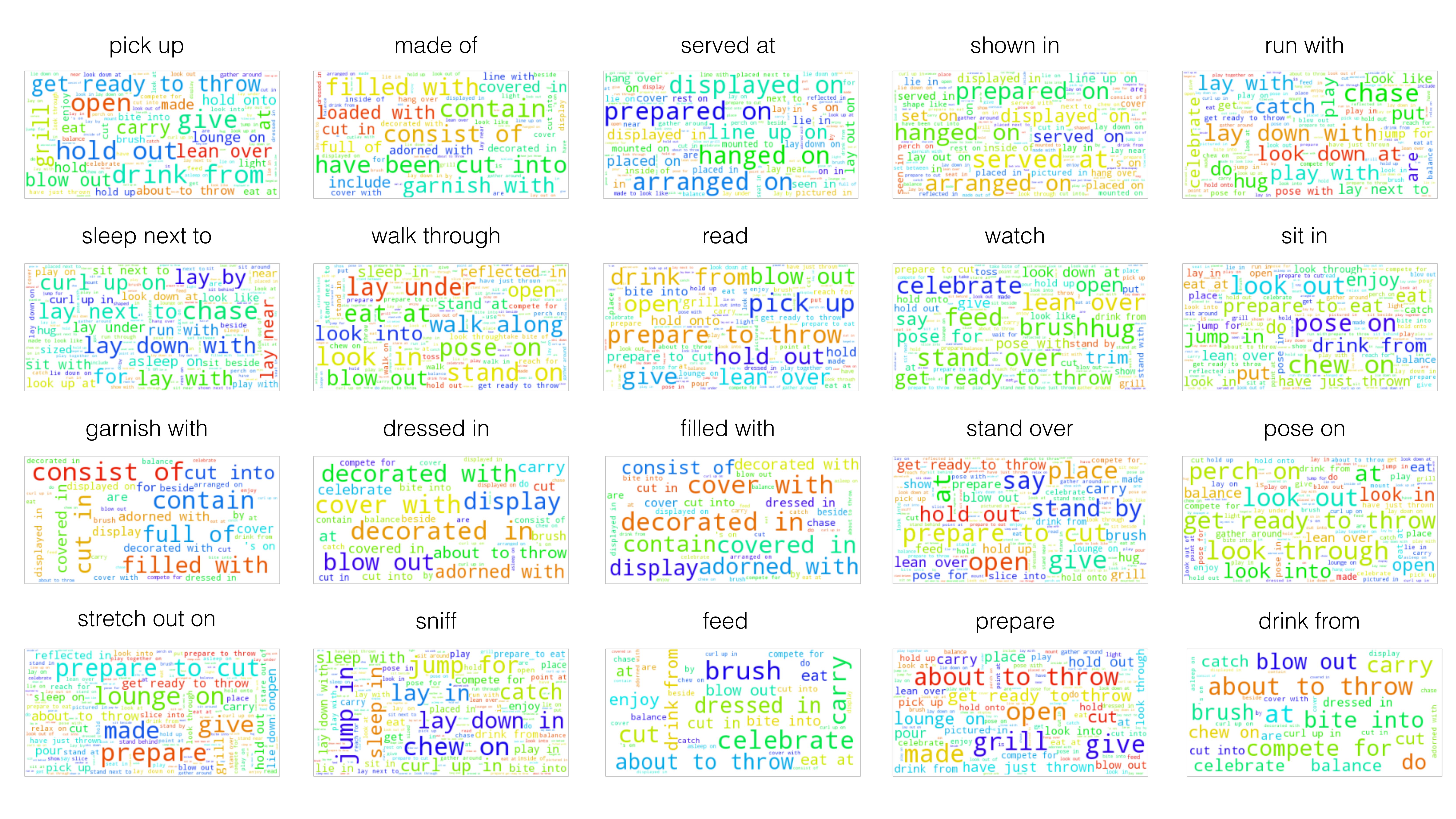}
\caption{Word cloud for a given relation indicates other relations co-occurring in the same cluster. Relations that co-occur more appear bigger than others. Observe how (visually) semantically close relations co-occur the most.}
\label{clusters}
\end{figure*}
\subsection{Cluster Visualizations}
We show the cluster visualizations for a randomly sampled set of relations from the CS VAL set (Fig.~\ref{clusters}). As in the main paper (Fig. 4), we analyze how frequently two relations co-occur in the same clusters. Interestingly, relations like \emph{drink from} co-occur with relations like blow out and bite into which all involve action with a person's mouth. 


\section{Visual Paraphrasing}
\label{sec:paraphrase}
The Visual Paraphrasing (VP) task~\cite{Lin_2015_CVPR} is to classify whether a pair of textual descriptions are paraphrases of each other. 
These descriptions have three sentence each. 
Table~\ref{tab:vp} presents results on VP for various settings of the model that are described below.

\compactparagraph{Model settings: }
We vary the number of hidden units $N_H \in \{50, 100, 200\}$ for both \vwv\coco{} and \vwv\wiki{} models.
We also vary our context window size to include entire description (\texttt{Descs}), individual sentences (\texttt{Sents}), window of size $5$ (\texttt{Winds}) and individual words (\texttt{Words}).
As described in~\refsec{subsec:vary-clusters}, we also have \texttt{Orig} and \texttt{PCA} settings.

\compactparagraph{Observations: }
From Table~\ref{tab:vp}, we see improvements over the text baseline~\cite{Lin_2015_CVPR}.
In general, \texttt{PCA} configuration outperforms \texttt{Orig} for low complexity models ($N_H=50$).
Using entire description or sentences as the context window gives almost the same gains, while performs drops when smaller context windows are used (\texttt{Winds} and \texttt{Words}).
As VP is a sentence level task where one needs to reason about the entire sentence to determine whether the given descriptions are paraphrases, these results are intuitive.

\section{Text-based Image Retrieval}
\label{sec:retrieval}
\begin{figure}
\includegraphics[width=\columnwidth]{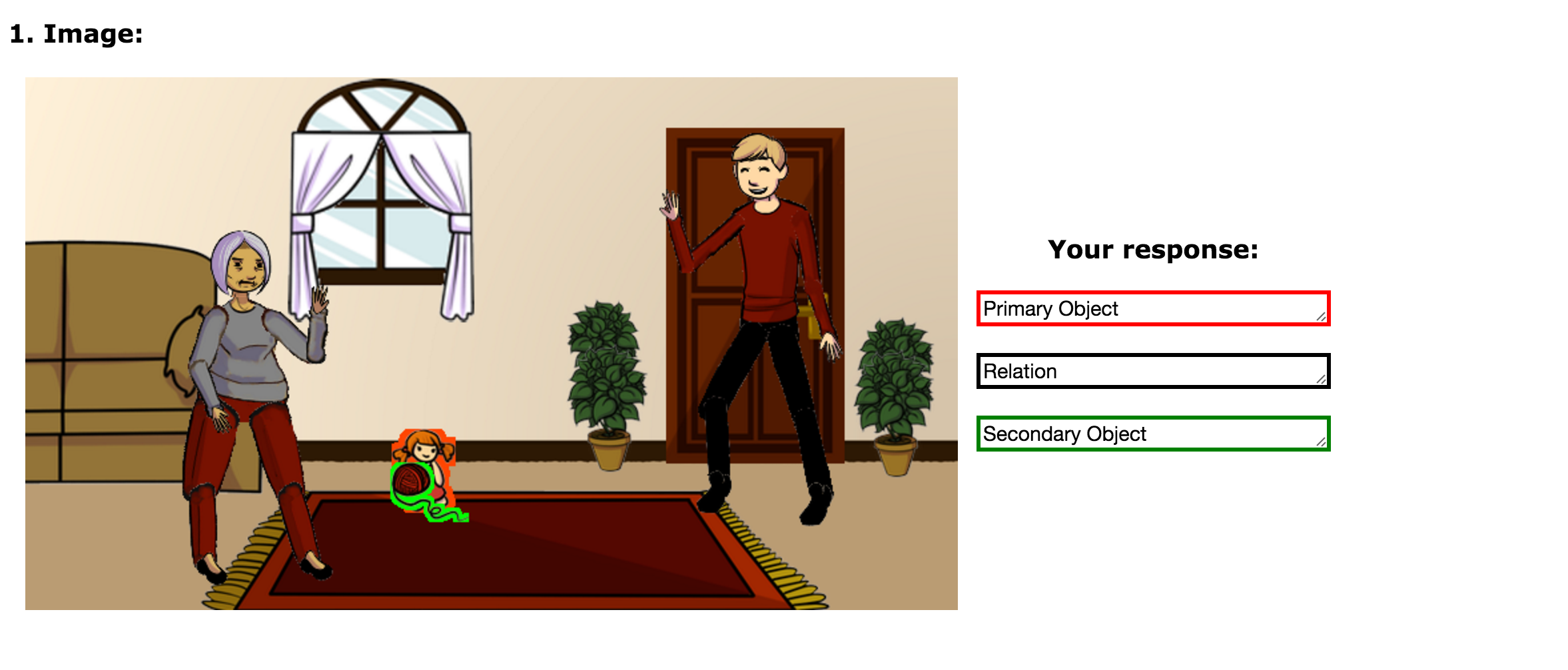}
\caption{An illustration of our tuple collection interface. Workers on AMT are shown the primary object (red) and secondary object (green) and asked to provide a tuple (Primary Object (P), Relation (R), Secondary Object (S)) describing the relation between them.}
\label{interface}
\end{figure}

Recall that in Text-based Image Retrieval (Sec.~4.3 in main paper), we highlight the primary  object (P) and secondary object (S) and ask workers on Amazon Mechanical Turk (AMT) to describe the relation illustrated by the scene with tuples. An illustration of our tuple collection interface can be found in Fig.~\ref{interface}. Each of the tuples entered in the text-boxes is treated as the query for text-based image retrieval.

\begin{figure}
\includegraphics[width=\columnwidth]{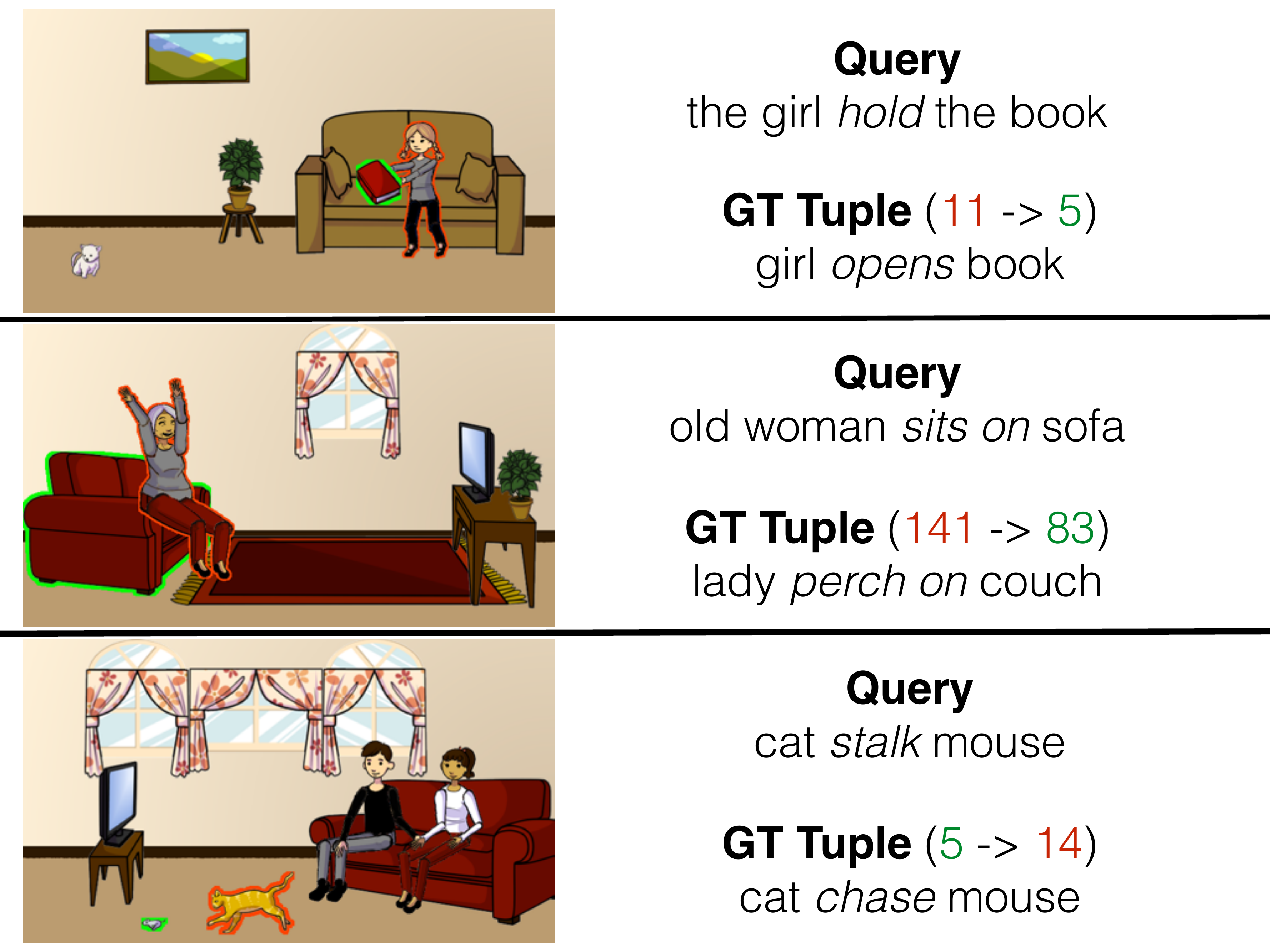}
\caption{We show qualitative examples for text-based image retrieval. We first show the query written by the workers on AMT for the image shown on the left. We then show the ground truth tuple and the rank assigned to it by \wv{} and then \vwv{} (\ie \wv{} $\rightarrow$ \vwv{}). The rank which is closer to the ground truth rank is shown in green. The first two examples are success cases, whereas the third shows a failure case for \vwv{}.}
\label{tbir_qualitative}
\end{figure}

Some qualitative examples of success and failure cases of \vwv{}\texttt{-wiki} with respect to \wv{}\texttt{-wiki} are shown in Fig.~\ref{tbir_qualitative}. We see that \vwv{}\texttt{-wiki} captures notions such as the relationship between \emph{holding} and \emph{opening} better than \wv{}\texttt{-wiki}.

\section{Real Image Experiments}
\label{sec:real-images}
We now present the results when training \vwv{} with real images from MSCOCO dataset by clustering using \texttt{fc7} features from the VGG-16~\cite{DBLP:journals/corr/SimonyanZ14a} CNN.

\compactparagraph{Intuition: }
We train \vwv{} embeddings with real images and compare them to those trained with abstract scenes, through the common sense task.

\compactparagraph{Setup: }
We experiment with two settings:
a) Considering all the $78k$ images from MSCOCO dataset, along with associated captions.
Each image has around $5$ captions giving us a total of around $390k$ captions to train.
We call \vwv{} trained on this dataset as \vwv\texttt{80k}.
b) We randomly select 213 relations from VAL set and collect 20 real images from MSCOCO and their corresponding tuples. 
This would give us $4260$ real images with tuples, depicting the 213  CS VAL relations. 
We refer to this model as \vwv\texttt{4k}.

We first train \vwv\texttt{80k} with $N_H = 200$ and use the \texttt{fc7} features as is, \ie. without PCA, in the \texttt{Sents} configuration (see \refsec{sec:paraphrase}).
Further, to investigate the complementarity between visual semantics learnt from real and visual scenes, we initialize \vwv\coco{} with \vwv\coco\texttt{80k}, \ie., we learn the visual semantics from the real scenes and train again to learn from abstract scenes.
Table~\ref{tab:real-images-80k} shows the results for \vwv\coco\texttt{80k}, varying the number of surrogate classes $K$.

We then learn \vwv\texttt{4k} with $N_H = 200$ in the \texttt{Orig/Words} setting (see \refsec{sec:commonsense}).
We observe that the performance on the validation set reduces for \vwv\coco\texttt{4k}.
Table~\ref{tab:real-images-4k} summarizes the results for \vwv\wiki\texttt{4k}.
\begin{table} \footnotesize
\setlength{\tabcolsep}{5.5pt}
\centering
\begin{tabular}{@{} c  c  c @{}}
\toprule
\multirow{2}{*}{$K$} & \multirow{2}{*}{\vwv\texttt{80k}} & \vwv\coco\\
&  & + \vwv\texttt{80k}\\
\midrule
$50$ & 73.6 & 74.7\\
$100$ & \textbf{73.7} & 74.5\\
$200$ & 73.4 & 74.2\\
$500$ & 73.2 & 73.8\\
$1000$ & 72.5 & \textbf{75.0}\\
$2000$ & 70.7 & 74.9\\
$5000$ & 68.8 & 74.6\\
\bottomrule
\end{tabular}
\caption{Performance on the common sense task of~\cite{vedantamLICCV15} using $78k$ real images with text baseline at $72.2$, initialized from \wv\coco.}
\label{tab:real-images-80k}
\end{table}

\begin{table} \footnotesize
\setlength{\tabcolsep}{5.5pt}
\centering
\begin{tabular}{@{} l  c c c c @{}}
\toprule
K & $25$ & $50$ & $75$ & $100$\\
\midrule
AP(\%) & $69.6$ & $70.6$ & $70.8$ & $70.9$\\
\bottomrule
\end{tabular}
\caption{Performance on the common sense task of~\cite{vedantamLICCV15} using $4k$ real images with with text baseline at $68.1$, initialized from \wv\wiki.}
\label{tab:real-images-4k}
\end{table}

\compactparagraph{Observations: }
From Table~\ref{tab:real-images-80k} and Table~\ref{tab:real-images-4k}, we see that there are indeed improvements over the text baseline of \wv{}.
The complementarity results (Table~\ref{tab:real-images-80k}) show that abstract scenes help us ground word embeddings through semantics complementary to those learnt from real images.
Comparing the improvements from real images (best AP of $73.7\%$) to those from abstract scenes (best AP of $74.8\%$), we see that that abstract visual features capture visual semantics better than real images for this task.
It if often difficult to capture localized semantics in the case of real images.
For instance, extracting semantic features of just the primary and secondary objects given a real image, is indeed a challenging detection problem in vision.
On the other hand, abstract scene offer these fine-grained semantics features therefore making them an ideal for visually grounding word embeddings.

\section{Comparison to Ren~\etal}
\label{sec:ren-et-al}
We next compare the embeddings from our \vwv{} model to those from Ren~\etal~\cite{xuimproving}. Similar to ours, their model can also be understood as a multi-modal extension of the Continuous Bag of Words (CBOW) architecture. More specifically, they use global-level \texttt{fc7} image features in addition to the local word context to estimate the probability of a word conditioned on its context.

We use their model to finetune word \wv\coco{} embeddings using real images from the MS COCO dataset. This performs slightly worse on common sense assertion classification than our corresponding (real image) model (Sec.~\ref{sub:real-im}) (73.4\% vs 73.7\%), while our best model gives a performance of 74.8\% when trained with abstract scenes. We then initialize the projection matrix in our \vwv{} model with the embeddings from Ren~\etal's model, and finetune with abstract scenes, following our regular training procedure. We find that the performance improves to 75.2\% for the \textbf{separate} model. This is a 0.4\% improvement over our best \vwv{} \textbf{separate} model. In contrast, using a curriculum of training with real image features and then with abstract scenes within our model yields a slightly lower improvement of 0.2\%. This indicates that the global visual features incorporated in the model of Ren~\etal, and the fine-grained visual features from abstract scenes in our model provide complementary benefits, and a combination yields richer embeddings.

\balance

{\small
\bibliographystyle{ieee}
\bibliography{egbib,rama,Mendeley}
}

\end{document}